\PassOptionsToPackage{table}{xcolor}
\PassOptionsToPackage{dvipsnames}{xcolor}
\documentclass[sigconf]{aamas} 

\usepackage{balance} 
\usepackage{config}
\DeclareUnicodeCharacter{2212}{-}

\makeatletter
\gdef\@copyrightpermission{
  \begin{minipage}{0.2\columnwidth}
   \href{https://creativecommons.org/licenses/by/4.0/}{\includegraphics[width=0.90\textwidth]{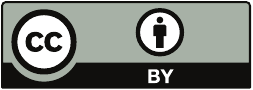}}
  \end{minipage}\hfill
  \begin{minipage}{0.8\columnwidth}
   \href{https://creativecommons.org/licenses/by/4.0/}{This work is licensed under a Creative Commons Attribution International 4.0 License.}
  \end{minipage}
  \vspace{5pt}
}
\makeatother

\setcopyright{none}
\acmConference[ALA '25]{Correspondence should to addressed to Tao Li (tl2636@nyu.edu). Preprint submitted to the Adaptive and Learning Agents Workshop (ALA 2025)}{May 19 -- 20, 2025}{Detroit, Michigan, USA, ala-workshop.github.io}{Avalos, Aydeniz, M\"uller, Mohammedalamen (eds.)}
\copyrightyear{2025}
\acmYear{2025}
\acmDOI{}
\acmPrice{}
\acmISBN{}
\title[Self-Confirming Transformer]{Self-Confirming Transformer for Belief-Conditioned Adaptation in Offline Multi-Agent Reinforcement Learning}

\author{Tao Li}
\affiliation{
  \institution{Department of Electrical and Computer Engineering,\\ New York University}
  \city{Brooklyn, NY}
  \country{United States}}
\email{tl2636@nyu.edu}

\author{Juan Guevara}
\affiliation{
  \institution{Department of Computer Science,\\ New York University Abu Dhabi}
  \city{ Abu Dhabi}
  \country{United Arab Emirates}}
\email{jdg8833@nyu.edu}

\author{Xinhong Xie}
\affiliation{
  \institution{Department of Computer Science and Engineering,\\ The Pennsylvania State University}
  \city{State College, PA}
  \country{United States}}
\email{xjx5116@psu.edu}

\author{Quanyan Zhu}
\affiliation{
  \institution{Department of Electrical and Computer Engineering,\\ New York University}
  \city{Brooklyn, NY}
  \country{United States}}
\email{qz494@nyu.edu}

\begin{abstract}
Offline reinforcement learning (RL) suffers from the distribution shift between the offline dataset and the online environment. In multi-agent RL (MARL), this distribution shift may arise from the nonstationary opponents in the online testing who display distinct behaviors from those recorded in the offline dataset. Hence, the key to the broader deployment of offline MARL is the online adaptation to nonstationary opponents. Recent advances in foundation models, e.g., large language models, have demonstrated the generalization ability of the transformer, an emerging neural network architecture, in sequence modeling, of which offline RL is a special case. One naturally wonders \textit{whether offline-trained transformer-based RL policies adapt to nonstationary opponents online}. We propose a novel auto-regressive training to equip transformer agents with online adaptability based on the idea of self-augmented pre-conditioning. The transformer agent first learns offline to predict the opponent's action based on past observations. When deployed online, such a fictitious opponent play, referred to as the belief, is fed back to the transformer, together with other environmental feedback, to generate future actions conditional on the belief. Motivated by self-confirming equilibrium in game theory, the training loss consists of belief consistency loss, requiring the beliefs to match the opponent's actual actions and best response loss, mandating the agent to behave optimally under the belief. We evaluate the online adaptability of the proposed self-confirming transformer (SCT) in a structured environment, iterated prisoner's dilemma games, to demonstrate SCT's belief consistency and equilibrium behaviors as well as more involved multi-particle environments to showcase its superior performance against nonstationary opponents over prior transformers and offline MARL baselines. 
\end{abstract}


\newcommand{\BibTeX}{\rm B\kern-.05em{\sc i\kern-.025em b}\kern-.08em\TeX}

\begin{document}


\pagestyle{fancy}
\fancyhead{}


\maketitle 

\section{Introduction}
Offline reinforcement learning (RL) has recently emerged as a promising alternative to online RL \cite{levin20off}, which extracts policies purely from the previously collected dataset without any interaction with the environment. As such, offline RL avoids online explorations required by online RL algorithms, which can be expensive (e.g., end-to-end robotic control \cite{e2e-rl}), dangerous (e.g., self-driving \cite{tao_cola}), and sometimes infeasible (e.g., healthcare \cite{rl-health}).  

Yet, a fundamental challenge of offline RL is the distribution shift between the offline training dataset and the online testing environment \cite{levin20off}. In plain words, the offline RL agent needs to properly handle unseen state-action pairs in the dataset during testing \cite{tao20causal}. When extending this offline RL  to multi-agent RL (MARL) settings, the distribution shift may be caused by exogenous agents who are beyond the preview of the trained MARL policy. We refer to these exogenous agents as the opponents. When opponents display a behavior pattern different from those included in the offline dataset, the ego agents are unprepared for these unseen state-action pairs resulting from opponents' unexpected moves. We refer to such an opponent as nonstationary, as it employs a different and possibly time-varying policy in testing, as opposed to the stationary policy used to collect offline data. As shown in one motivating example in \Cref{fig:madt-random}, blindly applying offline MARL policy gives degrading performance when playing with a nonstationary opponent.

While the study of developing autonomous agents capable of reasoning and adapting to unknown opponent policies, referred to as \textit{opponent modeling} (OM), is a long-standing and pivotal research topic in multi-agent systems and artificial intelligence \cite{stone18oppo-modeling}, this works explores the transformer architecture \cite{attention} to tackle opponent modeling and offline policy extraction combined as an integrated sequence modeling problem. The question we ask is 
\begin{flushleft}
       \textit{whether the offline transformer policy can adapt to the nonstationary opponent online by modeling its behavior auto-regressively}?  
\end{flushleft}

\begin{figure}
    \centering
    \includegraphics[width=1\linewidth]{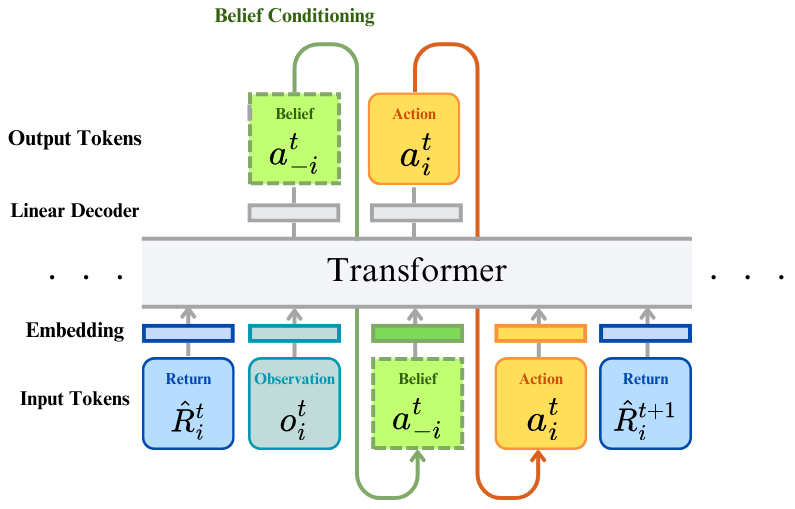}
    \caption{\footnotesize Self-augmented belief conditioning in the self-confirming transformer (SCT). SCT first generates a belief on the opponent’s action $a_{-i}^t$ (the green block), which is a fictitious token unobserved from the environment. Based on this belief, the transformer generates the action.  }
    \label{fig:sct}
    \Description{Self-Confirming Transformer (SCT) architecture. SCT first generates a conjecture on the opponent's action $a_{-i}^t$ (the green block) based on the hidden state $_{h}o_i^t$ of the observation $o_i^t$ produced by the transformer. This conjecture (or equivalently, its hidden state $_{h}a_{-i}^t$), together with the observation hidden state $_{h}o_i^t$, leads to the agent's action generation $a_i^t$ (the green arrow). The offline trained SCT enables the agent to reason its opponent's move in online testing. }
\end{figure}
This work answers this question affirmatively by introducing the self-confirming transformer (SCT) that learns to predict the opponent's move from the partial observation, which is then fed to the transformer itself to generate the ego agent's action, as depicted in \Cref{fig:sct}. The SCT is inspired by the self-confirming equilibrium (SCE) \cite{self-confirming}, a weaker variant of the seminal Nash equilibrium (NE) \cite{nash1951}, which does not assume every agent's compliance with NE policy. In contrast, SCE rests on subjective rationality, focusing on consistency between one's observations and subjective beliefs on the opponent's future move, which leads to the term ``self-confirming'' (see \Cref{def:sce}). From a sequence modeling perspective, the sheer difference between our SCT and previous transformers lies in that SCT creates a fictitious belief token as part of the input sequence, serving as the pre-conditioning for action generation, while prior works only use environment feedback (e.g., observations and rewards) to generate control actions. The additional belief generation in SCT, bearing a similar spirit to OM, prevents overfitting the offline data and helps the agent to adapt its action online. A summary of our contributions is as follows. 
\begin{itemize}
    \item We propose a novel auto-regressive self-confirming training, enabling the transformer to infer and adapt to the opponent's nonstationary policy without modifying its architecture. 
    \item We conduct extensive experiments in benchmark MARL environments to demonstrate SCT's greater adaptability over offline MARL baselines and recent transformer models. 
    \item We analyze SCT's equilibrium behavior in a structured iterated prisoner's dilemma, empirically certifying its self-confirming plays as instructed by self-confirming training. 
\end{itemize}
\section{Related Works}
\noindent\textbf{Offline Reinforcement Learning.} Offline RL methods extract policies from pre-collected datasets without interacting with the environment. These methods can be categorized into constraint-based and sequential model-based approaches \cite{meng23madt}. In constraint-based methods, off-policy algorithms treat offline datasets as a replay buffer to learn a policy with promising performance. However, the experiences in offline datasets and interactions with online environments have different distributions, i.e., there is a \textit{distribution shift}, causing overestimation in policy and value-based methods.  To address this issue, recent progress utilizes the conservatism idea \cite{levin20off} that compels the policy \cite{fujimoto2021minimalist} or value function estimation \cite{kumar2020conservative, fujimoto2019off} to the data manifold to control the extrapolation error.  

\noindent\textbf{Transformer in RL.} In addition to the constraint-based approach above, sequential model-based methods, which treat the offline policy training process as a sequence modeling problem, have also emerged as a powerful tool. Sequence modeling, motivated by the analysis of sequential data such as texts and time series, is concerned with learning the correlation among sequential data and forecasting or generating future data points, which has been the core research topic in natural language processing \cite{ilya14seq2seq}, speech recognition \cite{dong18speech-transformer}, and time-series prediction \cite{tao24picol,tao24dima}.
Since RL trajectories also display temporal correlations, sequence modeling methods aim to predict future states, actions, and rewards using past observations.

Due to transformers' encouraging success in sequence modeling \cite{attention}, recent efforts have been exploring the application of transformers in offline RL. One example of such is the decision transformer (DT) and its variants \cite{chen2021dt, furuta21generalDT}. DT learns the distribution of trajectories and predicts actions conditional on target rewards and previous observations. Instead of directly predicting the optimal actions,  the trajectory transformer and its variants \cite{levine21tt, wang22boot} use the transformer to roll out future trajectories and search for optimal policies. As for multi-agent RL, the current multi-agent transformer research mainly employs the transformer architecture for representation learning \cite{tao_multiRL, grover18encoder} under long horizons and partial observations, i.e., the transformer, learning the hidden representation of historical observations, becomes part of a bigger RL machinery for policy optimization \cite{yang22madt, meng23madt}. In contrast, our proposed transformer subscribes to DT and addresses the action generation conditional on the additional fictitious belief tokens. Another remark is that while those prior works on multi-agent transformers concentrate on cooperative tasks, a central transformer policy controls all agents; our approach alludes to independent learning \cite{claus_98jal,tao_info} where each agent handles the nonstationary opponents in noncooperative settings.    

\noindent\textbf{Opponent Modeling.} Our work is also closely related to transformer-based opponent modeling. The intuition is that modeling the opponent's behavior is also a sequence modeling problem, where the transformer learns to reconstruct opponent policies using offline data. Some early works utilize encode-decoder architecture (which is used by the transformer as well) to extract the temporal correlation between states and opponent actions \cite{grover18encoder, papoudakis21oppo}. Most relevant to our work are \cite{jing24ics, wallace24oppo-transformer}, where authors employ the transformer architecture to model the opponent's policy and forecast its actions for policy updates. These works treat the transformer as a purely predictive model for opponent action forecasting and call for additional mechanisms for policy learning. In contrast, our proposed SCT combines opponent action forecasting with policy generation using the vanilla decision transformer model \cite{chen2021dt}. We aim to investigate whether belief conditioning creates greater online adaptability when generating actions auto-regressively.   
\section{Preliminary}
\label{sec:pre}
\noindent\textbf{Multi-Agent Reinforcement Learning.} Consider learning in a multi-agent decision process described by a partially observable Markov game (POMG). A POMG with $N$ agents indexed by $i\in \{1,2,\dots, N\}:=[N]$ includes a global state space $\cS$, each agent's action space $\cA_i$, and a set of observations $\cO_i$ for each individual. The typical elements of these spaces are denoted by the corresponding uncapitalized letters. The time step is denoted by $t\in \N_{+}$, appearing as the superscript in the sequel. Unaware of the global state $s^t$, each agent receives a local observation $o^t_i\in \cO_i$ and chooses an action $a^t_i$. Denote by the bold symbol $\bm{o}^t=\{o_1^t,o^t_2,\ldots, o^t_N\}$ the joint observation. Then, with the joint actions of all agents, denoted by the bold symbol $\bm{a}^t=(a_1^t,a^t_2,\ldots, a^t_N)$, the environment transits to the next state $s^{t+1}$ according to the transition kernel $\cP: \cS\times \prod_{i\in [N]} \cA_i\rightarrow \Delta(\cS)$, and the decision-making process repeat. We assume all involved sets in this work are Borel sets (either discrete or continuous), and $\Delta(\cdot)$ denotes the Borel probability measure,  e.g., $\cP(s^{t+1}|s^t, \bm{a}^t)$ give the distribution of the next state. 

Agents' performance is evaluated through the reward function $r_i:\cS\times \prod_{i\in [N]} \cA_i \rightarrow \R$, and each agent aims to maximize its own discounted return $\sum_{t=1}^T \gamma^{t-1} r_i^t$, where $r_i^t=r_i(s^t, \bm{a}^t)$, and $T$ denotes the horizon length.  We consider the conventional decentralized information structure \cite{tao_info}, consisting of local feedback: $\cI^t_i=\{o_i^t, a_i^{t-1}, r_i^{t-1}\}$ in a non-cooperative multi-agent environment, including competitive and mixed cooperative-competitive scenarios \cite{lowe2017multiAC}, where agents may have distinct reward signals, i.e., $r_i\neq r_j$ for some $i,j\in [N]$. Each agent aims to find a policy $\pi_i\in \Pi_i$ that maps the past information to some action at each time step: $a_i^t\sim\pi_i(\cdot|\cI^{1:t}_i)$ to maximize the discounted return, where $\pi_i$ is assumed to be a stochastic policy yielding a distribution on $\cA_i$. We use $\pi_i$ and its neural network parameterization $\theta_i$ interchangeably to refer to the agent's policy. In general, agents' information structures are different, leading to the challenge of independent learning under asymmetric information and recursive reasoning \cite{hammer24col, tao24col}. This work explores the use of the transformer in sequence modeling to address independent learning under asymmetric observability. 

\noindent\textbf{Transformer Architecture.}
Generally, a transformer consists of an encoder, an attention module, and a decoder, which can use either the encoder, the decoder, or both, depending on the applications. Decoder-only models are useful for generating sequence and forecasting tasks \cite{gpt}. Encoder-only models are suitable for sequence understanding tasks \cite{devlin19bert}. We consider decoder-only architecture since the key of our SCT is to predict opponent actions. We now briefly review the attention module in the transformer model. 

The raw inputs of the transformer (which we will call tokens) are initially embedded in vectors of dimension $d_{model}$. Each input embedding generates a query, key, and value vector of dimensions $d_k$, $d_k$, and $d_v$. Vectors of the same type are stacked column-wise to produce three matrices $Q \in \mathbb{R}^{l \times d_k}$, $K \in \mathbb{R}^{l \times d_k}$, and $V \in \mathbb{R}^{l \times d_v}$, with $l$ the maximum context length (i.e., the length of the input sequence, defined as a hyperparameter).  The attention score is then calculated with the formula
\begin{equation*}
\text{Attention}(Q, K, V) = \text{softmax}\left({QK^\intercal}/{\sqrt{d_k}}\right) V.
\end{equation*}
The matrix $QK^\intercal$ is divided by $\sqrt{d_k}$ to prevent the vanishing gradient problem when applying the softmax row-wise \cite{attention}. After the softmax computation, {the upper off-diagonal triangle part of the resulting matrix  $\text{softmax}({QK^\intercal}/{\sqrt{d_k}})$ is masked with 0s}. This causal mask prevents future tokens from influencing the prediction of the current target and is the defining feature of a {causal transformer}, which is applied in our experiments since the opponent action prediction can only leverage historical observations. 

\noindent\textbf{Offline MARL as Sequence Modeling.} 
We take the multi-agent decision transformer (MADT) in \cite{meng23madt} as an example to illustrate the sequence modeling. Consider a trajectory $\tau$ from the offline dataset given by $\tau = \{ \boldsymbol{o}^1, \boldsymbol{a}^1,  \boldsymbol{o}^2, \boldsymbol{a}^2,  \ldots,  \boldsymbol{o}^T, \boldsymbol{a}^T \}$. MADT, parameterized by a decoder-only transformer model $q_\theta$, generates (predicts) sequential actions at each time step auto-regressively. Let $\hat{\tau}^t=\{ \boldsymbol{o}^1, \hat{\boldsymbol{a}}^1, \ldots,  \boldsymbol{o}^t, \hat{\boldsymbol{a}}^t \}$ be the truncated trajectory up to time $t$ with previous action predictions. Then, MADT's sequential generation proceeds as follows: $\hat{\boldsymbol{a}}^t=\argmax_{\boldsymbol{a}} q_\theta(\boldsymbol{a}|\hat{\tau}^{t-1},  \boldsymbol{o}^t)$. The learning objective of MADT is to minimize the distribution discrepancy between the generation $q_\theta$ and the offline data distribution $q_{\text{off}}$. Toward this end, one can consider the cross entropy (CE) loss to train MADT in discrete cases. Given predictions $\{\hat{\boldsymbol{a}}^t\}$, the CE loss is defined as $\cL_{CE}(\theta)=1/T \sum_{t=1}^T q_{\text{off}}(\hat{\boldsymbol{a}}^t)\log q_{\theta}(\hat{\boldsymbol{a}}^t|\hat{\tau}^{t-1},  \boldsymbol{o}^t)$.
While for continuous control tasks, the mean-squared error $\|\hat{\boldsymbol{a}}^t-\boldsymbol{a}^t\|^2$ leads to decent transformer policies as observed in \cite{chen2021dt}.  

Note that RL is a sequential decision-making process where the current actions influence future states and rewards. To equip the agent with forward-looking ability, DT slightly modifies the trajectory representation and adds the reward-to-go (return) $\hat{R}_i^t=\sum_{k=t}^T r_i^k$. The resulting trajectory is $\tau=\{\bm{\hat{R}}^1,\boldsymbol{o}^1, \hat{\boldsymbol{a}}^1, \ldots, \bm{\hat{R}}^t,  \boldsymbol{o}^t, \hat{\boldsymbol{a}}^t\}$, where the bold symbol $\bm{\hat{R}}^1$ denotes the agents' joint rewards, referred to as the return conditioning \cite{chen2021dt}. In plain words, such a return conditioning gives the transformer a sense of what to expect and directs its action generation to ensure that the cumulative rewards well approximate the return conditioning $\bm{\hat{R}}^1$. Such a practice is referred to as hindsight information matching (HIM), a popular technique in off-policy optimization \cite{furuta21generalDT}.

\section{Self-Confirming Transformer}
\begin{figure}[!ht]
\begin{subfigure}[t]{0.48\linewidth}
\centering
     \includegraphics[width=1\linewidth]{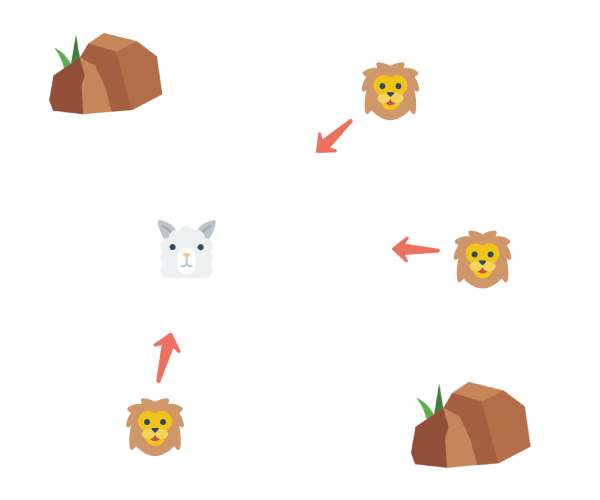}
    \caption{\footnotesize\texttt{simple-tag}: the three slow-moving predators aim to catch the fast-moving prey while avoiding the obstacles.}
    \label{fig:simple-tag}
\end{subfigure}
\hfill
\begin{subfigure}[t]{0.48\linewidth}
    \centering
     \includegraphics[width=1\linewidth]{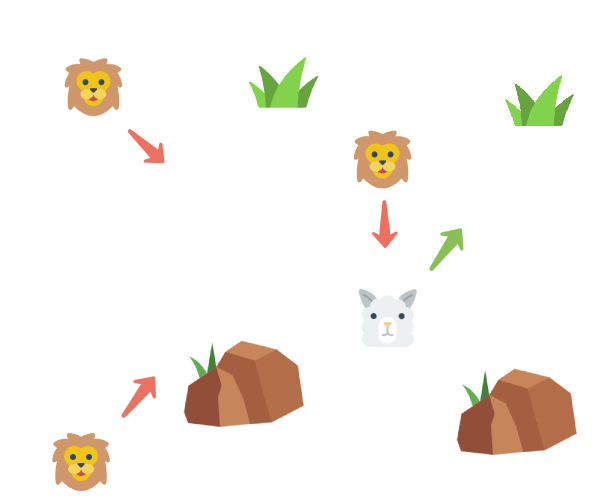}
    \caption{\footnotesize\texttt{simple-world}: a variant of \texttt{simple-tag} with two added food particles. The prey is rewarded when hitting the food. }
    \label{fig:simple-world}
\end{subfigure}
\caption{\footnotesize The predator-prey tasks in multi-agent particle environment. }
\Description{The illustration of two predator-prey tasks in the multi-agent particle environment: simple-tag and simple-world.  }
\end{figure}

\noindent\textbf{Motivating Example.} We consider the predator-prey task (a.k.a \texttt{simple-tag}) included in the multi-agent particle environment (MPE) \cite{lowe2017multiAC}, one of the benchmark environments in MARL. As shown in \Cref{fig:simple-tag}, the environment includes a prey who moves faster and aims to evade the three predators. The predators are slower and try to hit the prey while avoiding obstacles. 
\begin{figure}[!ht]
    \begin{subfigure}{0.48\linewidth}
         \includegraphics[width=1\linewidth]{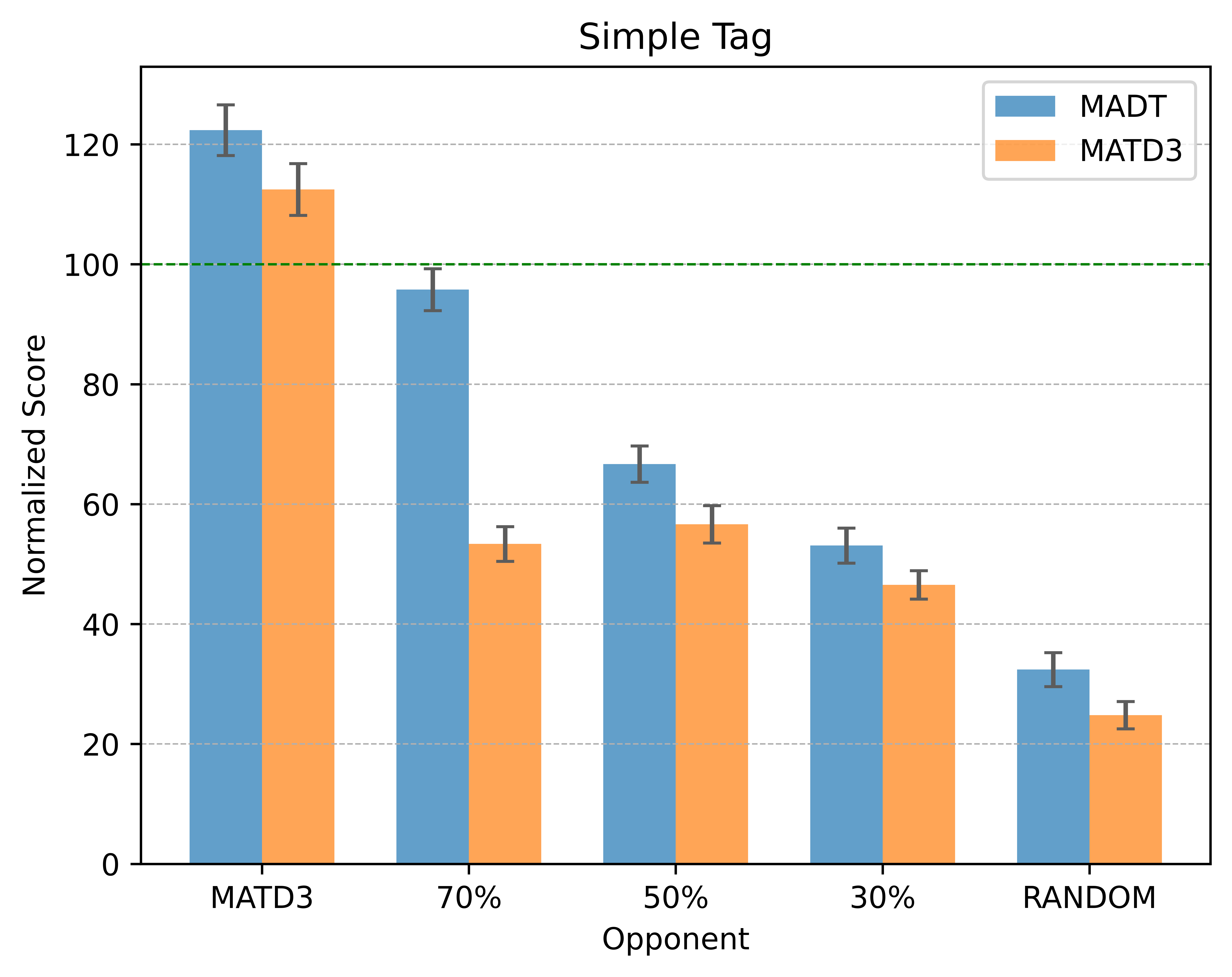}
    \end{subfigure}
    \hfill
   \begin{subfigure}{0.48\linewidth}
       \includegraphics[width=1\linewidth]{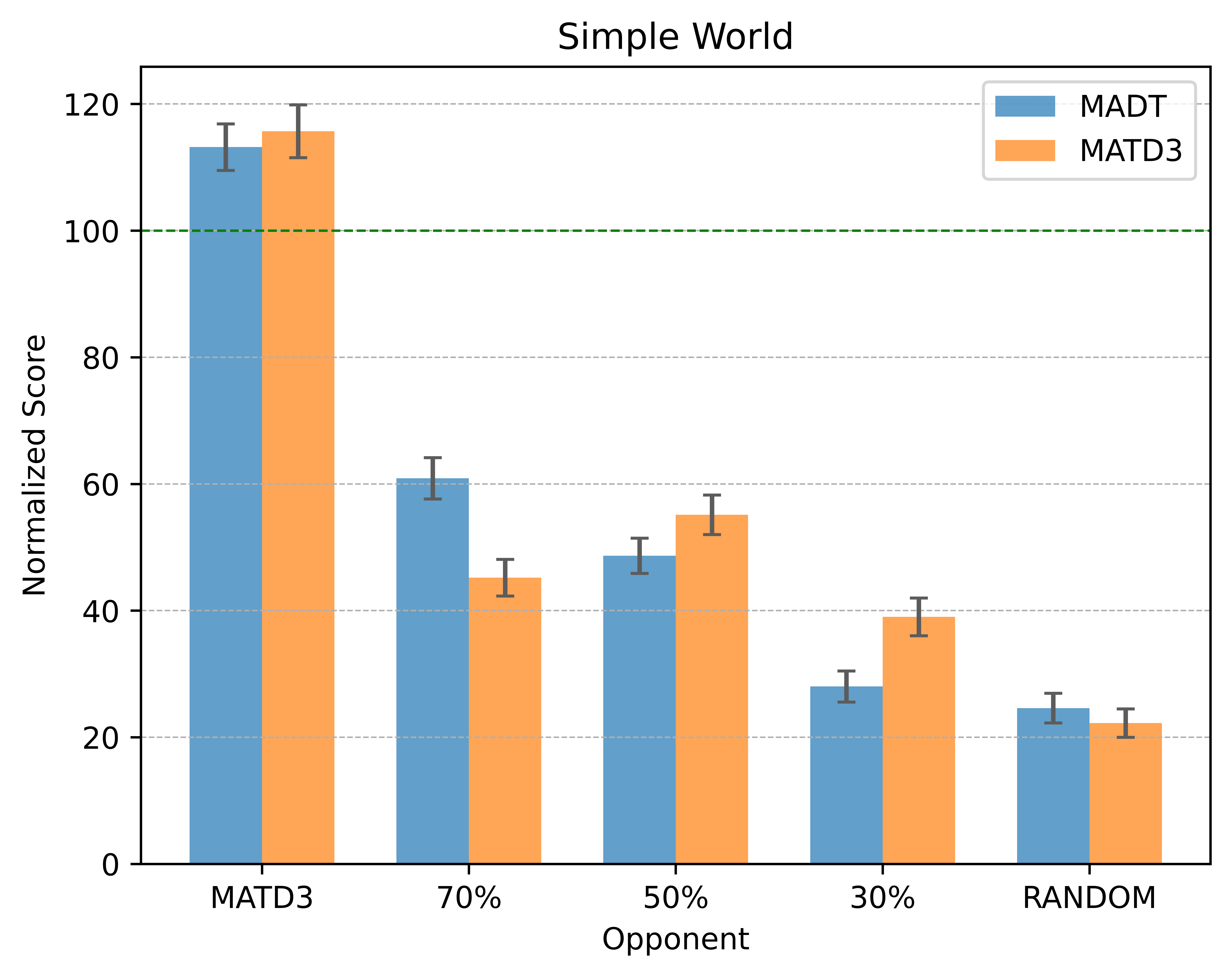}
   \end{subfigure}
    \caption{\footnotesize The normalized scores (the higher, the better) of playing MADT and  MATD3 policy against the nonstationary opponent in \texttt{simple-tag} (left) and \texttt{simple-world} (right). The opponent employs a blend of MATD3 and the random policy, with the blending rate $p$ shown on the x-axis. The green dashed line indicates the benchmark performance of the testing task.   }
    \label{fig:madt-random}
    \Description{The normalized scores (the higher, the better) of playing  MADT and  MATD3 policy against the nonstationary opponent in \texttt{simple-tag} (left) and \texttt{simple-world} (right). The opponent employs a blend of the benchmark MATD3 and the random policy, with the blending rate $p$ (shown on the x-axis) chosen from $\{1, 0.7, 0.5, 0.3, 0\}$. The smaller $p$ is, the more discrepant the blending policy is from the offline dataset. The green dashed line indicates the benchmark performance of the testing task. }
\end{figure}

The predators observe the relative positions and velocities of the prey, while the prey can only observe the relative positions of the other agents. All agents' actions are two-dimensional velocity vectors. Each time any one of the three predators collides with the prey, the former gets rewarded while the latter is penalized. The predator-prey is a mixed cooperative-competitive task, where the predators cooperate with each other to encircle the prey so that the rewards get tripled, while the game between the prey and predators is zero-sum like. Another environment we consider is \texttt{simple-world} shown in \Cref{fig:simple-world}, a more complicated variant of \texttt{simple-tag} as it includes two food particles that prey is rewarded for being close to. More details are deferred to \Cref{sec:exp}. 

We here briefly touch upon the training and the testing procedure, while the detailed experiment setup is included in \Cref{sec:exp}. We use MATD3 \cite{ackermann2019matd3}, one of the state-of-the-art MARL algorithms, to train the four agents (three predators and one prey) and collect expert-level trajectory data after MATD3 training stabilizes as the offline dataset. MADT is trained using offline trajectories of the three predators. The prey employs the following baseline policies during the testing: 1) $\pi^{M}$, the same MATD3 policy used to collect the data; 2) $\pi^{R}$, the random policy; 3) $\pi^B$, a blend of the random and the MATD3 policy. The random policy takes a uniform distribution over the action set regardless of the observation input. The blending policy works like a bang-bang controller: at each time step, the prey flips a coin first; if heads up, then it chooses $\pi^{M}$, otherwise $\pi^{R}$. This blend can be written as $\pi_{B}=p\times \pi^{M}+ (1-p)\times \pi^{R}$, where the parameter $p$ is the mean value of the binomial distribution, capturing the opponent's nonstationarity. 

The purpose of this example is to examine the MADT's online adaptability when facing a nonstationary opponent in testing. Since the opponent utilizes a policy distinct from that in training, the resulting trajectory deviates from the offline data. This adaptability concerns whether the MADT adjusts its action generation according to the changing trajectory distribution.  The evaluation metric is the normalized score, a customary metric indicating the discounted returns \cite{fu2020d4rl}. \Cref{fig:madt-random} reports the testing results, from which one can see that the MADT's performance gradually degrades as the opponent deviates from $\pi^{M}$.

Yet, one interesting phenomenon we observe is that the transformer-based policy does exhibit online adaptability compared with the pre-trained MARL policy, though to a limited extent. We equip the three predators with the MATD3 policies that are used in the data collection and let them play with the three baseline prey policies mentioned above. We denote the predators' MATD3 policies by $\pi_{pred}^M$. Note that $\pi_{pred}^{M}$ includes three MATD3 policies and one for each predator. \Cref{fig:madt-random} summarizes the testing results, from which one can see that $\pi_{pred}^M$ gives even lower scores than the MADT does (orange bars). We believe this adaptability originates from the generalization ability of the transformer architecture, which is also observed in large language models \cite{gpt, xie2024learning} and robotic transformers \cite{reed2022generalist}. This motivating example prompts one to ask whether offline-trained transformers can adapt to unseen opponents online. 

\noindent\textbf{Self-Confirming Belief Conditioning.} Reflecting on the return conditioning in MADT (see \Cref{sec:pre}), one realizes that such conditioning does not provide the agent direct contextual information regarding the opponent since the rewards $r_i^t$, which affects subsequent $\hat{R}_i^t$, are jointly determined by all agents' actions. Using the language of HIM \cite{furuta21generalDT}, the return conditioning fails to provide fine-grained \textit{information statistics} regarding nonstationary opponents.    

Taking inspiration from game-theoretic research on adaptive learning agents \cite{tao22confluence} and the self-confirming equilibrium \cite{self-confirming}, we propose to introduce an additional conditioning that represents the agent's subjective belief over the opponent's unobservable action. To articulate the intuition, we first digress from the transformer and briefly review the notion of self-confirming equilibrium (SCE). SCE was born from dissatisfaction with Nash equilibrium (NE) \cite{nash1951} since learning agents often display suboptimal and non-equilibrium behaviors \cite{self-confirming, pan-tao23noneq, pan-tao24variational}. As a relaxation to NE, SCE forgoes the \textit{collective rationality} that requires every agent's compliance with the NE policy. Instead, SCE embodies \textit{subjective rationality}, where the agent maximizes its rewards with respect to the beliefs over the opponent's policy, and the belief is consistent with observations. 

Mathematically, denote by $\cI_i^{1:t}$ the past observations up to time $t$. Similar to the agent's policy $\pi_i(\cdot|\cI_i^{1:t})\in \Delta(\mathcal{A}_i)$, the agent's belief is a distribution over the opponent's action set: $\mu_i(\cdot|\cI_i^{1:t})\in \Delta(\mathcal{A}_{-i})$. The definition of SCE in Markov games, adapted from its original version in extensive-form games \cite{self-confirming}, is as below. 
\begin{definition}[Self-Confirming Equilibrium]
\label{def:sce}
A strategy profile $(\pi_i, \pi_{-i})$ is a self-confirming equilibrium of the partially observable Markov game if, for each agent, there exists a belief $\mu_i$ such that 
    \begin{equation}
        \label{eq:sce-max}
        \pi_i\in \argmax_{\hat{\pi}_i\in \Pi_i} \E_{\hat{\pi}_i, \mu_i, \cP}\left[\sum_{t=1}^T \gamma^{t-1} r_i(s^t, a_i^t, a_{-i}^t)\right], 
    \end{equation}
where the belief $\mu_i$ is consistent with the opponent's equilibrium policy $\pi_{-i}$ with respect to all realizable information feedback in total variation, i.e., for all $\cI_i^{1:t}$, $\cI_{-i}^{1:t}$, $\mathbb{P}_{\pi_i, \pi_{-i}}[\cI_i^{1:t}, \cI_{-i}^{1:t}]>0, t=1,\ldots, T$, 
    \begin{equation}
    \label{eq:sce-consis}
        \sup_{A\subset \mathcal{A}_{-i}}\bigg|\mu_i(A|\cI_i^{1:t})-\pi_{-i}(A|\cI_{-i}^{1:t})\bigg|=0.
    \end{equation}
\end{definition}
Some remarks are in order. First, NE is a special case of SCE if the belief consistency in \eqref{eq:sce-consis} is imposed on every information structure \cite{self-confirming}, i.e., $\mu_i=\pi_{-i}$. Unlike NE, SCE does not mandate every agent to follow the optimal policy. It is likely that the opponent may employ arbitrary policies; as long as the ego agent can correctly identify these behavior patterns and best responds to the consistent belief, it still arrives at SCE. Second, the current practice of collecting offline datasets typically begins with centralized training of benchmark MARL algorithms, such as MADDPG \cite{lowe2017multiAC} and MATD3 \cite{ackermann2019matd3}, to learn optimal policies for each agent. When the training stabilizes, recording the sample trajectories in the replay buffer produces the desired dataset, referred to as the expert-level dataset. Essentially, the optimal policies with benchmark performance generate the recorded trajectories, with which one trains the transformer in the auto-regressive manner presented in \Cref{sec:pre}. Consequently, the transformer extracts the trajectory distribution under the NE policy and generates actions accordingly online, which could easily break down if the opponent does not follow the equilibrium policy \footnote{Even though, except for a few value-based algorithms \cite{hu03nashQ, amy03CEQ, Tao_blackwell} with certified equilibrium convergence, most policy-gradient-based algorithms prove to be convergent to stationary points or approximate equilibrium \cite{zhang2021multi, shutian23erm, pan-tao23meta-sg, pan-tao24delay, tao2024meta}, the key observation is that the resulting benchmark RL policies produce a fixed trajectory distribution corresponding to the stationary point in the offline dataset. Besides, the adapted SCE does not consider subgame perfectness under partial observability \cite{ouyang16dyna_asy, tao23pot} for simplicity.}.   

Even though our proposed transformer follows the standard decoder-only transformer architecture and typical offline training dataset that records equilibrium trajectory distribution, we aim to make the transformer agent mindful of the opponent's nonstationary behavior through an extra sequence modeling on the opponent's action, which bears the same spirit of the belief generation in SCE and opponent modeling (OM). What distinguishes ours from existing works on transformer-based OM is that we train the transformer in a self-confirming way so that the policy $\pi_i$ and belief $\mu_i$ are integrated into a single transformer, whereas prior works utilize transformers to represent $\mu_i$ and rely on additional policy search methods to configure $\pi_i$ \cite{papoudakis21oppo, jing24ics}.  

\noindent\textbf{Self-Confirming Loss.} Recall that the transformer model for agent $i$ is denoted by $q_{\theta_i}$, which generates actions auto-regressively using past local trajectory. Let $\hat{\tau}^t_i=\{\hat{R}_i^1, o_i^1, \hat{a}_i^1, \ldots, \hat{R}_i^t, o_i^t, \hat{a}_i^t\}$ be agent $i$ local trajectory (which only keeps individual information feedback from the trajectory $\tau$), and the action generation in vanilla DT is given by $\hat{a}_i^t=\argmax_{a\in \cA_i}q_{\theta_i}(a|\hat{\tau}_i^{t-1}, o_i^t)$.  

Inspired by \Cref{def:sce}, we first let the transformer predict the opponent's action [see \eqref{eq:sct-belief}], based on which the transformer generates the agent's action in responding to the belief [see \eqref{eq:sct-policy}].
\begin{subequations}
    \label{eq:sct}
    \begin{equation}
    \label{eq:sct-belief}
        \hat{a}_{-i}^t=\argmax_{a_{-i}\in \cA_{-i}}q_{\theta_i}(a_{-i}|\hat{\tau}^{t-1}, o_i^t),
    \end{equation}
    \begin{equation}
        \label{eq:sct-policy}
        \hat{a}_{i}^t=\argmax_{a_{i}\in \cA_{i}}q_{\theta_i}(a_{i}|\hat{\tau}^{t-1}, o_i^t, \hat{a}_{-i}^t).
    \end{equation}
\end{subequations}
Note that the opponent action $a_{-i}^t$ is unobservable to the ego agent during the implementation, and hence, the prediction $\hat{a}_{-i}^t$ is a fictitious token represents the agent's subjective inference extracted from the past observations. Such a token directs the transformer's future action generation, which we call belief conditioning. Compared with return conditioning, belief conditioning explicitly provides the information statistics of the opponent's actions. Of particular note is that such conditioning is created in a bootstrapping manner without intervention; that is, the transformer plays dual roles of both belief $\mu_i$ in \eqref{eq:sce-consis} and policy $\pi_i$ in \eqref{eq:sce-max}.    

After presenting its online implementation, we now shift the focus to offline training. To facilitate the discussion, we denote by $\pi_i$ and $\pi_{-i}$ (dropping the information feedback for simplicity) the policies used to collect the offline training data and assume $\pi_i$ is the best response to $\pi_{-i}$: $\pi_{i}\in\argmax_{\pi} \E_{\pi, \pi_{-i}}[\sum_{t=1}^T \gamma^{t-1}r_i^t]$. From \Cref{def:sce}, one can see that the transformer needs to ensure that 1) the belief generation is consistent with the opponent's actual action [see \eqref{eq:sce-consis}] and 2) the action generation is the best response to the belief [see \eqref{eq:sce-max}]. We propose the following cross-entropy loss: $\cL_\text{SCL}(\theta_i)=\cL_\text{belief}(\theta_i)+ \cL_\text{policy}(\theta_i)$, referred to as the self-confirming loss (SCL), to meet the two requirements simultaneously. 
\begin{subequations}
    \label{eq:scl}
    \begin{equation}
    \label{eq:scl-belief}
    \cL_\text{belief}(\theta_i)=1/T\sum_{t=1}^T\pi_{-i}(\hat{a}_{-i}^t)q_{\theta_i}(\hat{a}^t_{-i}|\hat{\tau}_i^{t-1}, o_i^t).
\end{equation}
\begin{equation}
    \label{eq:scl-policy}
    \cL_\text{policy}(\theta_i)=1/T\sum_{t=1}^T\pi_i(\hat{a}_{i}^t)q_{\theta_i}(\hat{a}^t_{i}|\hat{\tau}_i^{t-1}, o_i^t, \hat{a}_{-i}^t).
\end{equation}
\end{subequations}
Minimizing the belief consistency loss in \eqref{eq:scl-belief} is equivalent to minimizing the total variation distance between the SCT's belief generation and the opponent's actual policy since Pinsker's inequality tells that the square root of cross entropy upper bounds the total variation \cite{pinsker}. Ideally, the policy loss should correspond to the maximization problem in \eqref{eq:sce-max}. Yet, since we assume the offline policy $\pi_i$ is the best response, we can simply let the transformer imitate such policy by minimizing the cross entropy. The benefit is straightforward: two loss functions are cross entropy and fit the widely adopted auto-regressive training paradigm. 

Some remarks regarding the training practice are in order. First, if access to the offline policies $\pi_i$, $\pi_{-i}$ is not available, one can replace them with sample averages as in \cite{chen2021dt, furuta21generalDT}. Second, if the action is continuous, one can consider the mean-square loss \cite{chen2021dt}: $\cL_{SCL}(\theta_i)=\|a_{-i}^t-\hat{a}_{-i}^t\|^2+\|a_i^t-\hat{a}_i^t\|^2$, where $a_i^t$ and $a_{-i}^t$ denote the actions in the offline dataset, while $\hat{a}_i^t$ and $\hat{a}_{-i}^t$ are transformer's generated outputs. Third, the self-confirming loss does not include a weighing parameter to balance the two parts since the two are equally important. Finally, we remark that since the interdependency between the belief conditioning and the action generation, the SCL is actually a composite function, and its exact gradient computation $\nabla \cL_\text{SCL}(\theta_i)$ is sophisticated. Encouraged by the recent success of first-order gradient approximation in stochastic composite optimization \cite{nicol18reptile, tao2024meta}, we ignore the interdependency and compute the first-order gradient as if the belief and action generation were independent. 
\section{Experiments}
\label{sec:exp}
This section seeks to empirically answer the question we raised at the beginning: can SCT adapt to nonstationary opponents online? Relatedly, one may wonder whether the offline-trained belief generation produces consistent beliefs online. If so, to what extent, does SCT's success depend on the belief (ablation)? 

\noindent\textbf{Environments.} Following the benchmarking testbed in the literature, e.g., \cite{pan2022plan, tseng22knowledge}, we consider \texttt{simple-tag} and \texttt{simple-world} discussed in the motivating example. We begin with \texttt{simple-tag} and present the setup of the partial observation, action, and reward of each agent. The partial observation of the prey contains its own velocity and position, and its relative positions to obstacles and other agents. The action variable of the prey is a two-dimensional vector, each entry of which ranges from -1 to 1. As the prey aims to escape from predators, it gets a positive reward proportional (the factor is 0.1) to the sum of its distance from each predator, while it is penalized for being caught by any of the predators (-10 reward).  

The partial observation of one predator consists of its own velocity and position, its relative positions to the obstacles and other agents, and the prey's velocity. The predator's action space is the same as the prey's.   The predator is rewarded +10 after hitting the prey, otherwise penalized by the relative distance to the prey.  The obstacles are introduced to complicate the environment. Observable to all agents, obstacles are stationary once initialized within an episode. 
We set one prey, three predators, and two obstacles in this environment. Consequently, the observation space of prey is 14-dimensional: 2 for its velocity, 2 for its position, 4 for the relative position to obstacles, and 6 for other relative positions to agents. Similarly, the observation space of predators is 16-dimensional, and the additional two entries correspond to the prey's velocity.

\texttt{simple-world} is a more challenging task, where there are additional food particles that the prey is rewarded for being close to. the prey is rewarded +2 points for every time it hits a food particle. The environment includes only one obstacle. For the two environments, the episode length is $T=25$. Yet, when training the transformer models, the context length is $20$, i.e., the past 20 steps are used to calculate the loss. \Cref{tab:hyper} summarizes the hyperparameters involved in the training of SCT and baseline methods.
\begin{table}
\centering
\caption{\footnotesize  A summary of training hyperparameters.}
\label{tab:hyper}
\footnotesize
\begin{tabular}{ll}
\toprule
Hyperparameter         & Value                           \\
\midrule
\textbf{Self-Confirming Transformer } & \\
Maximum context Length & 20                              \\
Batch Size             & 64                              \\
Hidden Dimensions      & 128                             \\
\# of Layers           & 3                               \\
\# of Attention Heads  & 1                               \\
Activation function~   & ReLU                            \\
\# Steps per epoch     & 10000                           \\
\multirow{2}{*}{\# Epochs} & 1 for Medium and Expert\\
                        &  10 for  Random   \\
Learning Rate          & 1e-4                            \\
Weight Decay           & 1e-4                 \\
\hline
\textbf{Behavior Cloning} & \\
Maximum Context Length & 20    \\
Batch Size             & 64    \\
Hidden Dimensions      & 128   \\
\# of Layers           & 3     \\
Dropout                & 0.1   \\
\# Steps per Epoch     & 10000 \\
\# Epochs~                & 15    \\
Learning Rate          & 1e-4  \\
Weight Decay           & 1e-4 \\ 
\bottomrule
\end{tabular}
\end{table}

\noindent\textbf{Offline Datasets.} The offline trajectories representing random, medium, and expert levels of play are divided into three datasets, where each dataset consists of 1 million transitions. The random dataset is obtained from unrolling episodes of a randomly initialized policy. The medium dataset is obtained by stopping the training phase of MATD3  once it reaches a medium level of play and then unrolling the episodes. The expert-level dataset is given by collecting transitions from the MATD3 once it is fully trained. 

\noindent\textbf{Baselines.} We conduct a comparative study between SCT and existing works based on imitation learning, offline MARL, and sequence modeling. Specifically, we consider the following baselines.

\noindent\textbf{1) Behavior Cloning (BC).} Behavior Cloning is an imitation learning algorithm \cite{hussein17imitate-survey} where the three predators' behavior recorded in the dataset is replicated. The implementation consists of a Multilayer Perceptron with ReLu activation and dropout. The input consists of the three predator's observation histories concatenated and flattened. We utilize mean squared error loss during training, with the dataset's actions as ground truth. The hyperparameters are summarized in \Cref{tab:hyper}.

\noindent\textbf{2) Multi-Agent Batch-Constrained Q-Learning (MA-BCQ).}  Batch-constrained Q-learning \cite{fujimoto2019off} imposes constraints on the action space to compel the agent to align more closely with on-policy behavior regarding a subset of the provided data. We implement MA-BCQ (\textbf{MA-BCQ}) based on the BCQ implementation provided by \cite{Yang2021BelieveWY}. Considering the fact that BCQ employs two Q networks for a single agent, MA-BCQ includes six Q networks, as each predator needs two critics. The QMixer network in MA-BCQ takes in six Q values and outputs one Q value to evaluate the joint actions of predators. We follow the hyperparameter setup in \cite{fujimoto2019off}. 

\noindent\textbf{3) Offline MARL with Actor Rectification (OMAR).} Assuming an actor-critic architecture, OMAR uses zeroth-order information to rectify the critic so as to update the actor conservatively. In addition to BCQ and OMAR, there exist many other competitive baselines, such as CQL \cite{kumar2020conservative} and ICQ \cite{Yang2021BelieveWY}. However, it is reported in \cite{pan2022plan} that OMAR outperforms CQL and ICQ in \texttt{simple-tag} and \texttt{simple-world}. We follow the official implementation of OMAR offered by the authors \cite{pan2022plan}.

\begin{figure}
    \begin{subfigure}{\linewidth}
        \includegraphics[width=1\linewidth]{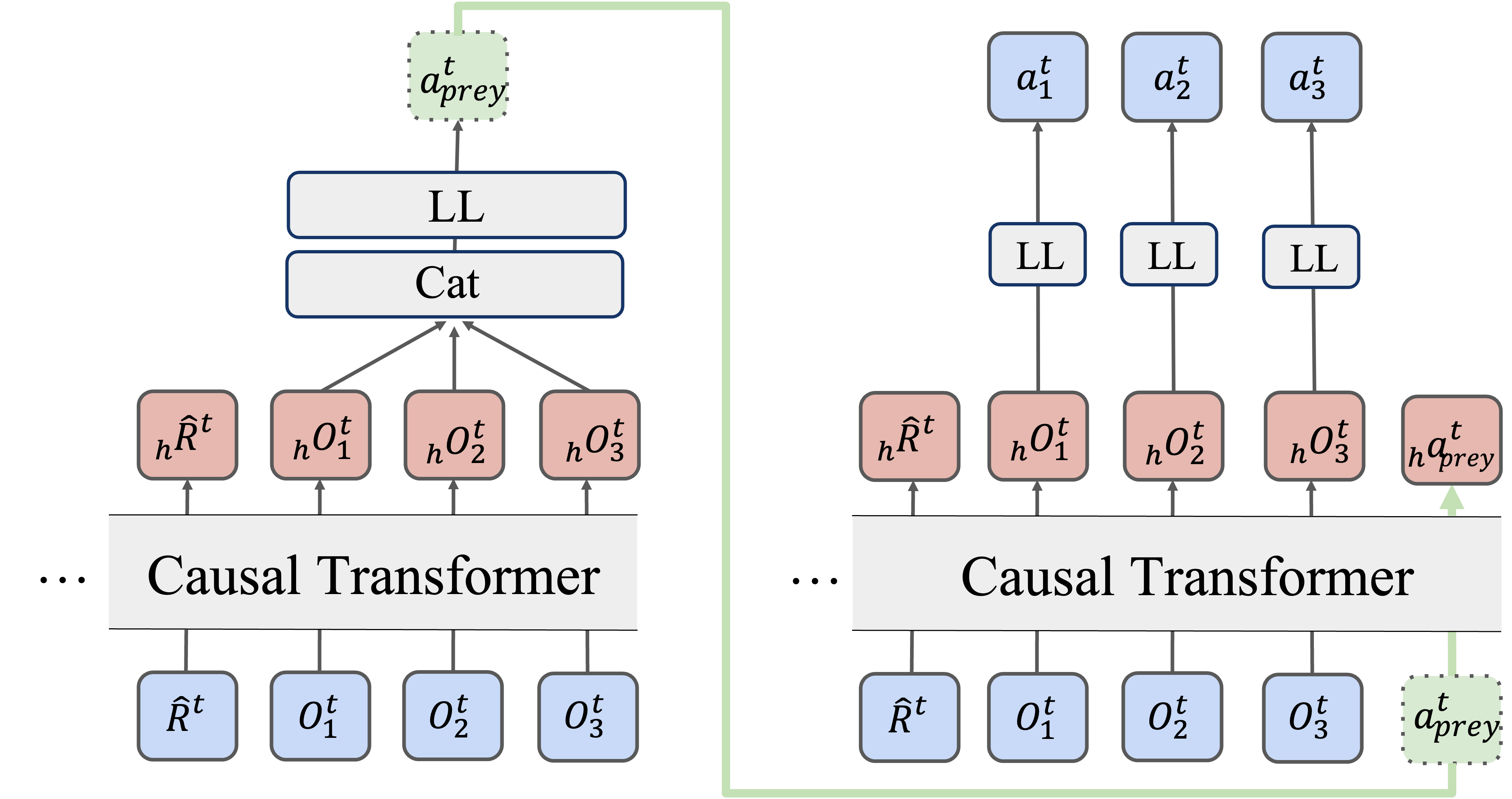}
    \caption{\footnotesize RMADT operation. }
    \label{fig:rmadt-arch}
    \end{subfigure}
    \begin{subfigure}{\linewidth}
    \includegraphics[width=1\linewidth]{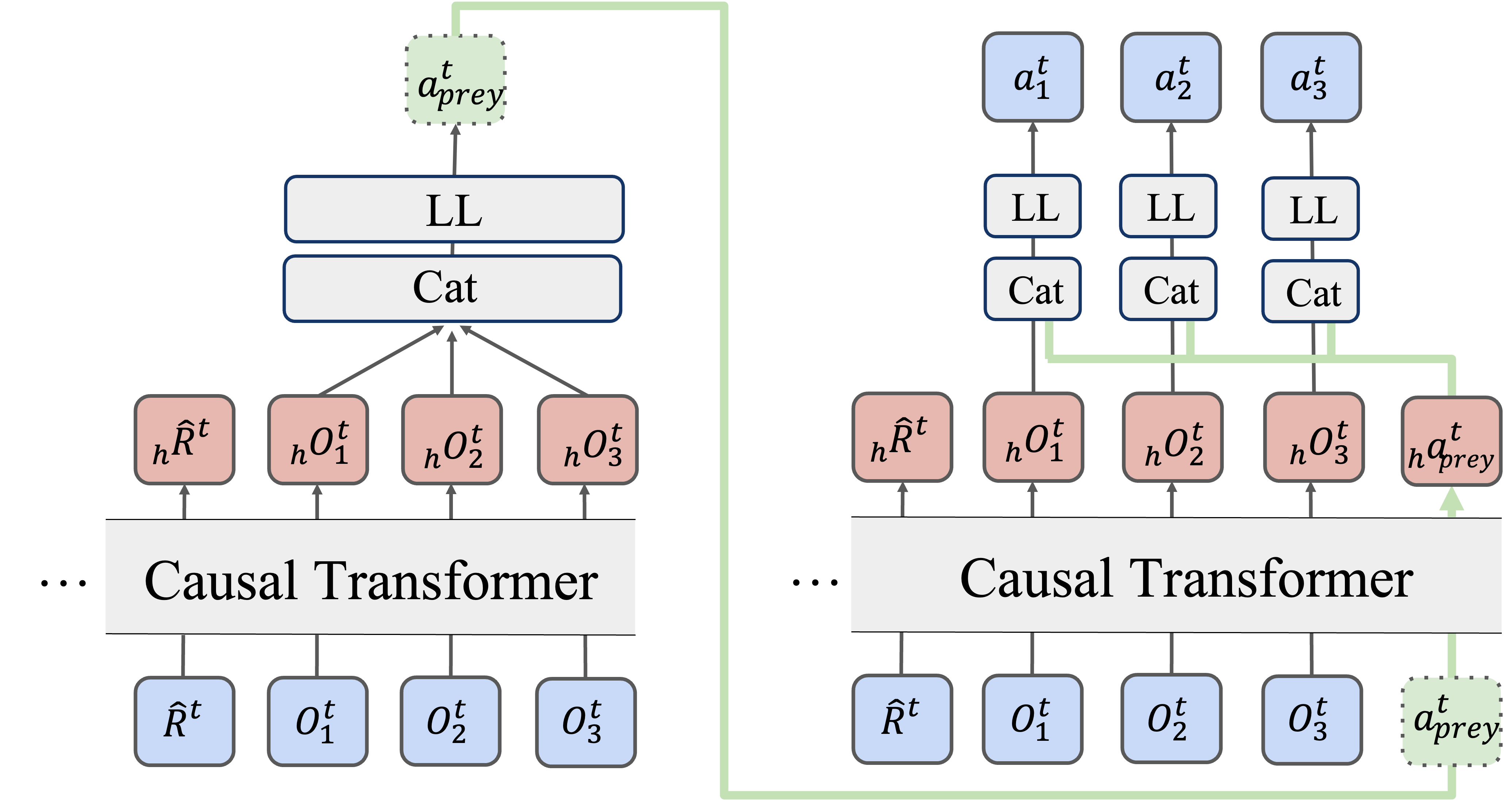}
    \caption{\footnotesize SCT operation.}
    \label{fig:sct-arch}
    \end{subfigure}
    \caption{\footnotesize A comparison between RMADT and SCT operation. The belief generation in RMADT does not direct the action generation. Even though the two share the same loss function, RMADT only aims to accurately predict the opponent's action, and the resulting action is not self-confirming.}
    \label{fig:transformer-compare}
    \vspace{-0.5cm}
    \Description{A comparison between RMADT and SCT architecture. The belief generation in RMADT does not direct the action generation. Even though the two share the same loss function, RMADT only aims to accurately predict the opponent's action, and the resulting action is not self-confirming.}
\end{figure}

\noindent\textbf{4) Transformer Models.} Finally, we consider transformer-based models for the ablation studies, which include the \textbf{MADT} discussed in the motivating example. To investigate the role of belief conditioning, we consider a middle point between MADT and SCT, which we call belief-regularized MADT (RMADT). Similar to SCT, RMADT also generates a belief using past observations, yet such a belief is not fed back to the transformer for action generation. It only appears in the belief loss as a regularizer to adjust the auto-regressive training. \Cref{fig:transformer-compare} visualizes the RMADT and SCT operation. In SCT's multi-agent implementation, the transformer (attention module) first generates the hidden states $\prescript{}{h}{o_1}^t, \prescript{}{h}{o_2}^t, \prescript{}{h}{o_3}^t$ of the three predators' partial observations, which are then concatenated (Cat) and passed through a linear layer (LL) to predict the opponent's action (the green block), which becomes part of the input stream. In contrast, RMADT merely generates the prey's action prediction and aims to close the gap between the belief and the actual opponent's action. Yet, the belief is not used for action generation.  

\noindent\textbf{Opponent in Testing.}  To evaluate the adaptability of the predators, the prey is controlled by five distinct policies for each task. These opponent policies include 1) the \textbf{MATD3} policy, the one used to collect the training data, 2) \textbf{MADDPG} policy, an actor-critic policy trained for each environment, 3) \textbf{Random} policy: a heuristic-based policy designed to randomly sample feasible actions, 4) \textbf{Still} policy: a simple policy that freezes the prey at the initialized location,  5) \textbf{Blend} policy, the blending policy introduced in the motivating example with $50\%$ blending rate. 

\begin{figure}[!ht]
\begin{subfigure}{0.48\linewidth}
    \includegraphics[width=1\textwidth]{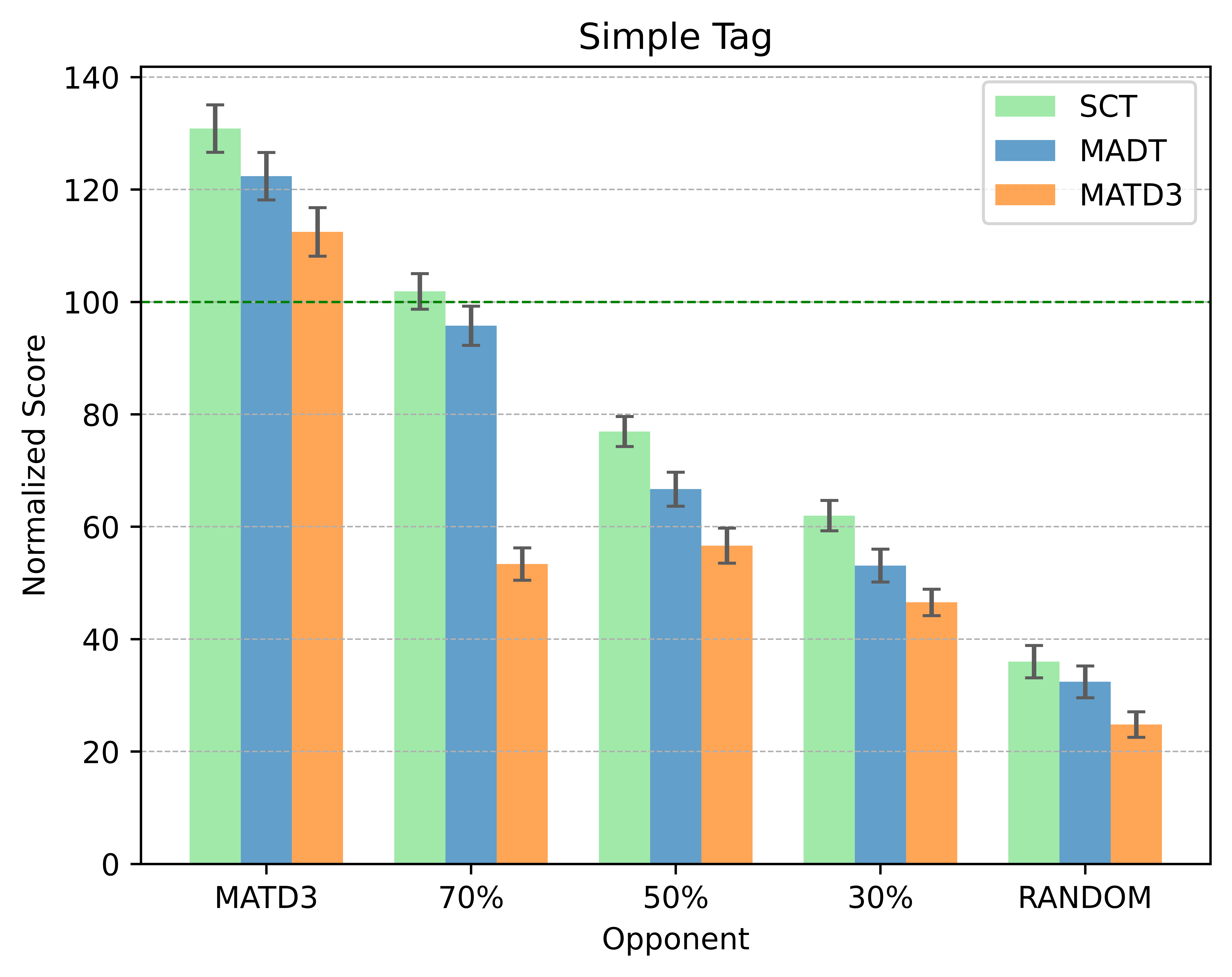}
\end{subfigure}
\hfill
\begin{subfigure}{0.48\linewidth}
    \includegraphics[width=1\textwidth]{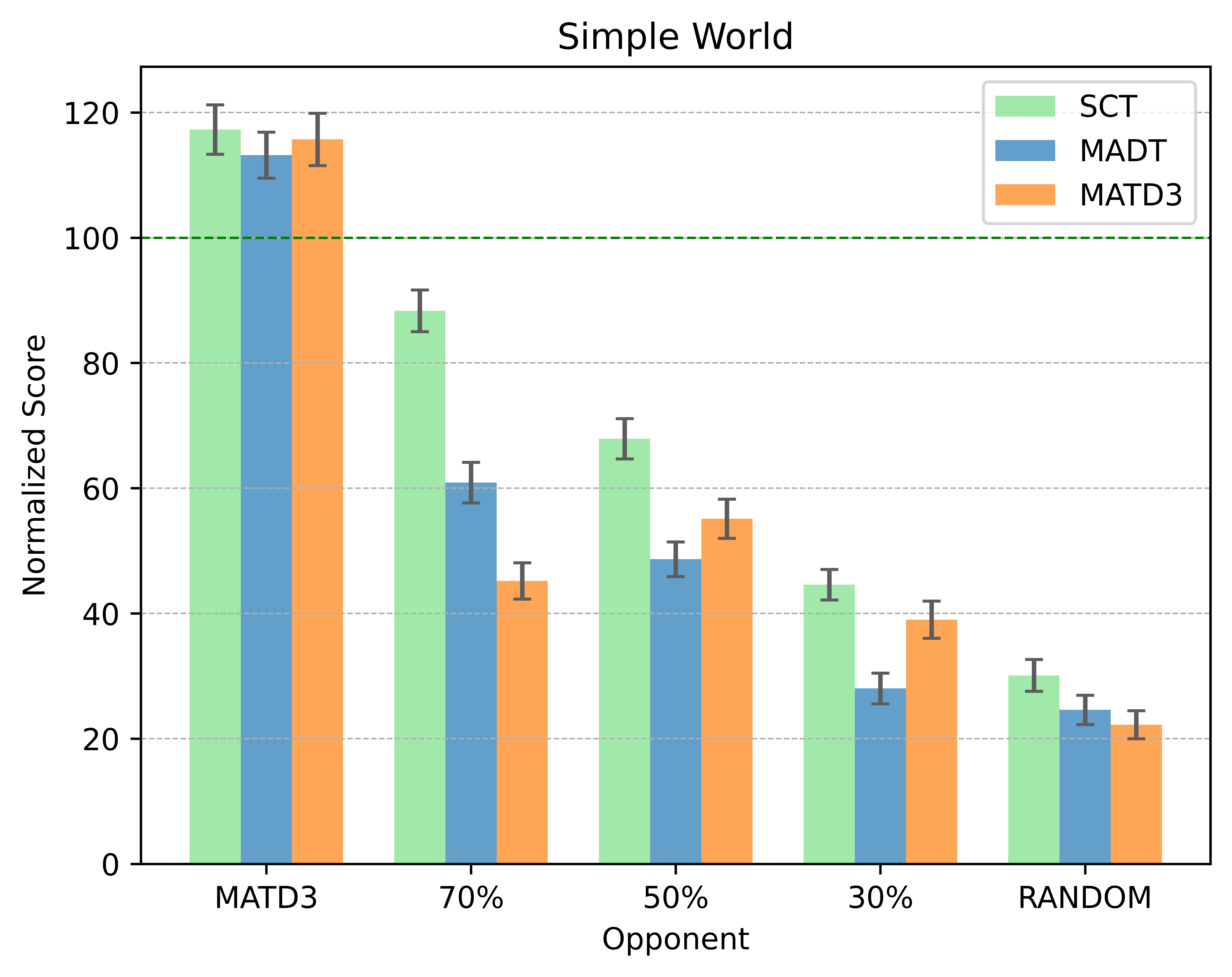}
\end{subfigure}
\caption{\footnotesize The normalized scores of SCT in \texttt{simple-tag} and \texttt{simple-world} environments. SCT outperforms MADT when facing nonstationary opponents.}
\label{fig:sct-curse}

\Description{A comparison of normalized scores of SCT in simple-tag and simple-world environments.}
\end{figure}

\noindent\textbf{Quantitative Results.} First, to complete the story in the motivating example, we add SCT's normalized scores to the bar plots in \Cref{fig:madt-random}, leading to \Cref{fig:sct-curse}. The figure suggests that SCT adapts better to nonstationary opponents in testing than MADT.

We also report in \Cref{tab:simple-tag} the mean and standard deviation of normalized scores (see \cite[Sec. 5]{yu2021benchmarking}) based on 100 runs using different random seeds in \texttt{simple-tag} and \texttt{simple-world}.  We observe that SCT consistently outperforms both MA-BCQ and OMAR across all experiments, except for the random dataset case where MA-BCQ leads extensively. Furthermore, SCT's performance is on par with or superior to the BC approach. Notably, SCT exhibits greater adaptability compared to its basic counterpart, as indicated by higher mean rewards in most experiments (orange entries in \Cref{tab:simple-tag}). We also observe that most of the outliers (colored in red) in \Cref{tab:simple-tag} pertain to MA-BCQ pre-trained over the random dataset, for which we speculate that the batch constraint better controls the extrapolation error than OMAR. Since this work focuses on transformer models, we leave the speculation for future investigation.  
\begin{table}
\caption{\footnotesize The normalized scores of SCT and baseline algorithms trained under the expert, medium, and random datasets in \texttt{simple-tag} and \texttt{simple-world}. SCT exhibits greater online adaptability than those baselines. }
\label{tab:simple-tag}
\footnotesize
\resizebox{\columnwidth}{!}{
\footnotesize
\begin{tabular}{ccccccc}
\toprule
                            &  Simple-Tag    & MATD3                 & MADDPG               & Still              & Random                & Blend                \\
\midrule
\multirow{6}{*}{\rotatebox{90}{Expert}}  & OMAR       & $103.19\pm 8.29$          & $9.05\pm 1.62$          & $38.80\pm 7.54$          & $24.13\pm 4.08$           & $58.02\pm 4.95$          \\
                                     & BC         & $121.11\pm 7.81$          & $11.22\pm 3.24$          & $44.57\pm 6.8$           & $35.7\pm 4.95$            & $73.65\pm 5.04$          \\
                                     & MA-BCQ     & $113.92\pm 8.16$          & $10.52\pm 1.52$          & $31.67\pm 6.83$          & $31.26\pm 5.25$           & $67.39\pm 5.97$          \\
                                     & MADT       & $123.32\pm 8.04$          & $8.16\pm 1.57$          & $44.10\pm 7.9$           & $32.83\pm 5.48$           & $76.90\pm 5.23$          \\
                                     & RMADT      & $122.94 \pm 4.50$         & $7.53\pm 1.80$         & $32.52\pm 2.74$          & $38.28 \pm 4.29$          & $70.04 \pm 20.87$        \\
                                     & SCT        &  $\cellcolor{Dandelion}\mathbf{126.20\pm 7.48}$ & $\cellcolor{Dandelion}\mathbf{11.94\pm 1.79}$ & $\cellcolor{Dandelion}\mathbf{54.87\pm 7.54}$ & $\cellcolor{Dandelion}\mathbf{38.98\pm 4.97}$  & $\cellcolor{Dandelion}\cellcolor{Dandelion}\mathbf{92. 87\pm 5.92}$ \\
\midrule
\multirow{6}{*}{\rotatebox{90}{Medium}}& OMAR       & $72.61\pm 6.69$           & $10.78\pm 1.21$          & $42.49\pm 6.67$          & $24.68\pm 3.73$           & $44.14\pm 4.78$          \\
                                     & BC         & $77.49\pm 6.93$           & $11.60\pm 2.72$          & $43.20\pm 6.39$          & $31.41\pm 4.11$           & $54.05\pm 4.35$          \\
                                     & MA-BCQ     & $56.08\pm 6.05$           & $10.36\pm 1.39$          & $33.75\pm 6.27$          & $25.65\pm 3.98$           & $49.51\pm 4.6$           \\
                                     & MADT       & $73.96\pm 5.76$           & $11.98\pm 1.60$          & $37.81\pm 5.96$          & $27.58\pm 4.05$           & $50.47\pm 4.42$          \\
                                     & RMADT      & $74.67\pm 3.5$            & $8.90\pm 1.31$         & $29.21\pm 1.93$          & $30.53 \pm 2.2$ & $44.15 \pm 2.10$         \\
                                     & SCT        & $\cellcolor{Dandelion}\mathbf{79.33\pm 5.80}$  & $\cellcolor{Dandelion}\mathbf{12.22\pm 1.55}$ & $\cellcolor{Dandelion}\mathbf{52.87\pm 6.02}$ & $\cellcolor{Dandelion}\mathbf{34.78\pm 3.52}$           & $\cellcolor{Dandelion}\mathbf{61.54\pm 5.05}$ \\
\midrule
\multirow{6}{*}{\rotatebox{90}{Random}}& OMAR       & $6.71\pm 3.03$            & $-1.48\pm 0.43$           & $1.53\pm 2.06$           & $1.50\pm 1.11$            & $3.53\pm 1.51$           \\
                                     & BC         & $−0.31\pm 1.13$           & $-3.27\pm 0.94$          & $-1.48\pm 0.86$          & $-1.48\pm 1.09$           & $0.03\pm 1.08$           \\
                                    & MA-BCQ     & $\cellcolor{RedOrange!60}\mathbf{32.86\pm 5.73}$  & $\cellcolor{RedOrange!60}\mathbf{3.79\pm 1.16}$  & $\cellcolor{RedOrange!60}\mathbf{28.1\pm 6.94}$  & $\cellcolor{RedOrange!60}\mathbf{6.18\pm 2.94}$   & $\cellcolor{RedOrange!60}\mathbf{12.9\pm 3.16}$  \\
                                     & MADT       & $9.94\pm 2.56$            & $-0.23\pm 0.63$           & $4.31\pm 2.01$           & $3.07\pm 1.49$            & $4.97\pm 1.65$           \\
                                     & RMADT      & $8.76\pm 1.36$            & $-2.43\pm 0.20$           & $3.44 \pm 1.23$          & $0.71 \pm 4.25$           & $4.93 \pm 0.88$          \\
                                     & SCT        & $24.75\pm 2.73$           & $-0.03\pm 0.78$           & $6.36\pm 1.73$           & $5.03\pm 1.51$            & $6.37\pm 1.57$   \\
\toprule
                            &  Simple-World    & MATD3                 & MADDPG               & Still              & Random                & Blend                \\
\midrule
\multirow{6}{*}{\rotatebox{90}{Expert}} & OMAR         & $114.26\pm 3.52$ & $-4.29\pm 8.33$ & $34.52\pm 3.76$ & $23.58\pm 2.21$ & $48.33\pm 2.82$ \\
                                     & BC           & $111.08\pm 3.12$ & $-9.05\pm 6.10$ & $37.73\pm 3.68$ & $25.31\pm 2.16$ & $53.14\pm 2.52$ \\
                                     & MA-BCQ       & $106.59\pm 3.38$ & $-2.75\pm 6.79$ & $45.33\pm 2.96$ & $21.62\pm 1.63$ & $51.07\pm 2.81$ \\
                                     & MADT         & $105.83\pm 3.18$ & $-2.85\pm 6.76$ & $31.69\pm 3.33$ & $23.34\pm 21.3$ & $48.61\pm 2.66$ \\
                                     & RMADT        & $110.44\pm 3.42$ & $-1.54\pm 7.70$ & $41.76\pm 3.98$ & $23.69\pm 2.13$ & $75.90\pm 3.58$ \\
                                    & SCT          & $\cellcolor{Dandelion}\mathbf{115.20\pm 3.38}$ & $\cellcolor{Dandelion}\mathbf{-1.32\pm 7.15}$ & $\cellcolor{Dandelion}\mathbf{53.31\pm 3.39}$ & $\cellcolor{Dandelion}\mathbf{28.28\pm 2.29}$ & $\cellcolor{Dandelion}\mathbf{92.87\pm 5.92}$ \\
\midrule
\multirow{6}{*}{\rotatebox{90}{Medium}} & OMAR         & $73.81\pm 4.46$  & $-0.84\pm 7.23$ & $\cellcolor{RedOrange!60}\mathbf{58.67\pm 3.73}$ & $31.37\pm 1.73$ & $42.23\pm 2.55$ \\
                                     & BC           & $86.23\pm 2.39$  & $-5.48\pm 10.47$ & $41.11\pm 3.38$ & $29.84\pm 2.06$ & $43.45\pm 2.32$ \\
                                     & MA-BCQ       & $76.99\pm 3.31$  & $-3.00\pm 9.88$ & $36.02\pm 3.71$ & $30.12\pm 1.99$ & $40.92\pm 2.38$ \\
                                     & MADT         & $81.70\pm 2.84$  & $0.54\pm 6.51$ & $28.92\pm 2.82$ & $26.07\pm 1.95$ & $43.32\pm 2.22$ \\
                                     & RMADT        & $86.87\pm 2.93$  & $0.96\pm 8.11$ & $30.25\pm 3.79$ & $28.52\pm 2.13$ & $55.78\pm 2.92$ \\
                                    & SCT          & $\cellcolor{Dandelion}\mathbf{87.13\pm 2.97}$  & $\cellcolor{Dandelion}\mathbf{1.30\pm 5.51}$ & $33.41\pm 3.02$ & $\cellcolor{Dandelion}\mathbf{31.87\pm 1.96}$ & $\cellcolor{Dandelion}\mathbf{57.87\pm 2.66}$ \\
\midrule
\multirow{6}{*}{\rotatebox{90}{Random}} & OMAR         & $8.37\pm 1.16$   & $-5.51\pm 0.87$  & $4.54\pm 1.01$  & $5.41\pm 0.77$  & $6.39\pm 0.85$  \\
                                     & BC           & $−0.62\pm 0.62$  & $-13.41\pm 5.28$ & $-0.75\pm 0.58$ & $0.06\pm 0.69$  & $0.15\pm 0.68$  \\
                                     & MA-BCQ       & $6.52\pm 1.42$   & $\cellcolor{RedOrange!60}\mathbf{-3.11\pm 7.98}$  & $\cellcolor{RedOrange!60}\mathbf{6.59\pm 1.38}$  & $3.40\pm 0.81$  & $4.01\pm 0.85$  \\
                                     & MADT         & $5.28\pm 1.10$   & $-7.21\pm 5.87$  & $1.27\pm 0.96$  & $4.85\pm 0.89$  & $4.28\pm 0.80$  \\
                                     & RMADT        & $8.36\pm 1.27$   & $-7.54\pm 5.96$  & $2.93\pm 1.01$  & $4.61\pm 0.82$  & $\cellcolor{RedOrange!60}\mathbf{8.12\pm 0.93}$  \\
                                    & SCT          & $\cellcolor{Dandelion}\mathbf{8.95\pm 1.42}$   & $-4.47\pm 5.80$  & $4.93\pm 0.81$  & $\cellcolor{Dandelion}\mathbf{6.09\pm 0.91}$  & $6.37\pm 1.57$  \\
\bottomrule
\end{tabular}
}
\end{table}

\noindent\textbf{Ablation.}  We compare SCT with MADT and RMADT to see to what extent the belief generation contributes to the SCT's success. Recall that MADT is trained to generate predators' actions using the offline trajectories without any beliefs about the prey. Even though RMADT also generates the belief and uses the same loss function as SCT, i.e., $\cL(\theta_i)=\|\hat{a}^t_{prey}-a_{prey}^t\|^2+\|\hat{a}^t_{pred}-a_{pred}^t\|^2$. However, the belief $\hat{a}^t_{prey}$ and the action $\hat{a}^t_{pred}$ simultaneously based on past observations, whereas SCT first generates the conjecture that later serves as the input to the action generation, see \Cref{fig:transformer-compare} for visualization of SCT and RMADT. We record the opponent's action prediction accuracy of SCT and RMADT in \texttt{simple-tag} and \texttt{simple-world}. The accuracy metric is defined as follows.  
\begin{equation*}
    \textbf{Accuracy} = \frac{\# \text{steps with accurate predictions} }{\# \text{total steps}}.
\end{equation*}
We consider an opponent's action prediction $\hat{a}_{-i}^t$ accurate if its relative error falls within an $\epsilon$-neighborhood of the ground truth:  $\hat{a}_{-i}^t\in \{a: \|a-a_{-i}^t\|/\|a_{-i}^t\|<\epsilon\}$. The prediction accuracy of an episode indicates the number of steps at which the prediction is accurate. We set the radius to be $\epsilon=10\%$. As shown in \Cref{tab:oppo-act}, SCT and RMADT return comparable results, suggesting that the two acquire similar forecasting abilities. Hence, the superiority of SCT, as indicated in \Cref{tab:simple-tag}, shows that belief conditioning in SCT plays a bigger part than regularization in RMADT.   
\begin{table}
    \caption{\footnotesize A comparison of the prediction accuracy of SCT and RMADT in \texttt{simple-tag} and \texttt{simple-world}.}
    \label{tab:oppo-act}
    \centering
    \resizebox{\columnwidth}{!}{
    \begin{tabular}{ccccccc}
    \toprule
                                   & simple-tag & MATD3 & MADDPG & Still & Random & Blend \\
    \midrule
\multirow{2}{*}{\rotatebox{90}{Exp}}  & SCT      & 1.000 & 0.690  & 0.978 & 0.310  & 0.803 \\
                                   & RMADT    & 1.000 & 0.611  & 0.910 & 0.293  & 0.690 \\
    \midrule
\multirow{2}{*}{\rotatebox{90}{Med}}  & SCT      & 1.000 & 0.665 & 0.996 & 0.300  & 0.888 \\
                                   & RMADT    & 1.000 & 0.644  & 0.988 & 0.324  & 0.872 \\
    \midrule
\multirow{2}{*}{\rotatebox{90}{Rand}} & SCT      & 1.000 & 0.823  & 1.000 & 0.301  & 0.865 \\
                                   & RMADT    & 1.000 & 0.843  & 1.000 & 0.300  & 0.630 \\
\toprule
                                  & simple-world & MATD3 & MADDPG & Still & Random & Blend \\
\midrule
\multirow{2}{*}{\rotatebox{90}{Exp}} & SCT      & 1.000 & 0.082  & 1.000 & 0.285  & 0.713 \\
                                  & RMADT    & 1.000 & 0.088  & 1.000 & 0.282  & 0.661 \\
\midrule
\multirow{2}{*}{\rotatebox{90}{Med}} & SCT      & 1.000 & 0.102  & 1.000 & 0.293  & 0.650 \\
                                  & RMADT    & 1.000 & 0.088  & 1.000 & 0.229  & 0.577 \\
\midrule
\multirow{2}{*}{\rotatebox{90}{Rand}} & SCT      & 1.000 & 0.092  & 1.000 & 0.285  & 0.707 \\
                                  & RMADT    & 1.000 & 0.054  & 1.000 & 0.287  & 0.627 \\
\bottomrule
    \end{tabular}
}
\end{table}

\noindent\textbf{Self-Confirming Play in Iterated Prisoner's Dilemma.}
We further apply the proposed SCT to the iterated prisoner's dilemma (IPD) \cite{mathieu2017}. Compared with more sophisticated MARL environments above, IPD presents a repeated matrix game environment, enabling a close inspection of the equilibrium behaviors of the proposed SCT. The primary question we ask is: does SCT play the SCE? Similar to our previous discussions, we need to investigate 1) whether the SCT can form accurate beliefs on the opponent's play and 2) whether the SCT can optimize its own action sequence based on its beliefs. More importantly, the SCE in IPD admits a simple strategy representation, which is referred to \textit{win-stay-lose-shift} (WSLS) \cite{nowak93win-stay} or \textit{pavlov} \cite{milinski96pavlov}, to be introduced later with the game matrix. We aim to inspect if SCT's action generation coincides with SCE.

\noindent\textbf{Iterated Prisoner's Dilemma.} We begin by introducing the prisoner's dilemma game. Suppose two criminals (the row and column players) are arrested and imprisoned, each of whom can cooperate (denoted by C) for mutual benefit (a lesser charge for both, denoted by $R$) or betray their partner (``defect'', denoted by D) for individual freedom (denoted by $T$) while leaving the other with the charge (denoted by $S$). If they both choose to defect, then the two face the same charge (denoted by $P$). \Cref{tab:ipd} presents the payoffs of players under different action profiles. A customary setup requires that $T>R>P>S$ and $\text{T}+\text{S}< 2\text{R}$, and the classical chosen values are presented in the table \cite{mathieu2017}. In IPD, two players play the matrix game in \Cref{tab:ipd} repeatedly with full observations of the other's past plays. The iterated plays enable players to adjust their strategies to the opponent's behavior pattern. One of the self-confirming equilibria in IPD is given by the \textbf{pavlov} strategy: cooperates on the first move and then, starting from the second round, cooperates if and only if both players opt for the same action in the previous move. 
\begin{table}
\centering
\caption{\footnotesize The payoff matrix of the prisoner dilemma.}
\label{tab:ipd}
\begin{tblr}{
  cell{1}{3} = {c=2}{},
  cell{3}{1} = {r=2}{},
  vline{3-5} = {2,4}{},
  vline{2-5} = {3}{},
  hline{2} = {3-4}{},
  hline{3-5} = {2-4}{},
  vline{2} = {4-5}{}
}
          &           & Player II &        \\
          &           & (C)ooperate & (D)efect \\
Player ~I & (C)ooperate & R=3, R=3      &S=0, T=5   \\
          & (D)efect    & T=5, S=0      & P=1, P=1   
\end{tblr}
\end{table}

\noindent\textbf{Offline Dataset.} We curate a training dataset by collecting action sequences under diverse strategies provided in \cite{mathieu2017}. The selected strategies include  \textbf{all\_d}, \textbf{all\_c}, \textbf{tit\_for\_tat}, \textbf{spiteful}, \textbf{soft\_majo}, \textbf{hard\_majo}, \textbf{per\_ccd}, \textbf{per\_ddc}, \textbf{mistrust}, \textbf{per\_cd}, \textbf{tf2t}, \textbf{hard\_tft}, \textbf{slow\_tft},   \textbf{gradual}, \textbf{prober}, and \textbf{mem2}. The detailed descriptions of these strategies are in \cite[Sec. 3.3]{mathieu2017}. Each strategy participates in a round-robin tournament against all other strategies, including itself. Each game consisted of 100 rounds, and the results were recorded in the final dataset.  SCT is trained on this data to predict the opponent's action and use this prediction to generate its own. The loss function follows \eqref{eq:scl}.

\noindent\textbf{Evaluation.} We test {SCT}'s adaptability to different opponents' strategies by playing 
the transformer against each strategy recorded in the training dataset. We compare the normalized cumulative payoffs and prediction accuracy (the higher, the better) and present the results in \Cref{tab:sct-ipd-seen}. Note that when generating beliefs and actions, we configure the transformer to pick the one with the maximum likelihood, and hence, there is no stochasticity in testing (unlike the previous experiments). The final cumulative reward was 4840 (unnormalized for ranking purposes), placing {SCT} as the best strategy overall (refer to \cite{mathieu2017} for the complete ranking). Moreover, we measure its prediction accuracy against each opponent and obtain a mean accuracy of $97.82\%$.

In addition, we inspect {SCT}'s adaptation ability to unseen strategies by testing the transformer against a new family of strategies. We consider a class of finite-memory-based strategies defined in \cite{mathieu2017}, denoted by $\textsc{memory}(X, Y)$, which determines the future actions using the ego agent's last $X$ plays and opponent's $Y$ plays. Following the definition in \cite{mathieu2017},  $\textsc{memory}(X, Y)$ begins with the $\max(X, Y)$ first moves, and the agent observes the $\max(X, Y)$ rounds of prisoner dilemma game. Then, based on its last $X$ actions and its opponent's last $Y$ actions, the agent determines the future plays using a deterministic mapping from $\{\text{C}, \text{D}\}^{X+Y}$ to $\{\text{C}, \text{D}\}$. We pick ten best-performing $\textsc{memory}(1, 2)$ and $\textsc{memory}(2, 1)$ strategies in the tournament ranking \cite[Sec. 5.10]{mathieu2017} and  refer to these strategies, in order, as $\textsc{mem}_s$, for $1\leq s\leq 10$. 

We report the reward and belief accuracy in \Cref{tab:mem-testing-result} when testing SCT against these unseen memory-based strategies.  The total reward is 2752, with a mean accuracy of $90.9\%$ when predicting these strategies. This places the {SCT} as the best strategy in the tournament with the unseen strategies, surpassing the second by a margin of 200. 
\begin{table}[]\
    \centering
     \caption{\footnotesize Normalized rewards and prediction accuracy of SCT against training strategies.}
    \label{tab:sct-ipd-seen}
    \resizebox{\linewidth}{!}{
    \begin{tabular}{ccccccccc}
    \toprule
                        & \textbf{all\_d} & \textbf{tit\_for\_tat} & \textbf{spiteful} & \textbf{soft\_majo} & \textbf{hard\_majo} & \textbf{per\_ddc} & \textbf{per\_ccd} & \textbf{mistrust} \\
                        \midrule
      \textbf{reward}   &  87.93 & 110.25 & 142.34 & 107.59 & 144.81 & 118.60 & 116.57 & 143.36 \\ 
      \textbf{accuracy} & 0.99 & 0.99 & 0.99 & 0.99 & 0.98 & 0.96 & 0.98 & 0.98 \\ 
      \toprule
      \textbf{all\_c} & \textbf{per\_cd} & \textbf{pavlov} & \textbf{tf2t} & \textbf{hard\_tft} & \textbf{slow\_tft} & \textbf{gradual} & \textbf{prober} & \textbf{mem2} \\ 
    \midrule
    122.40 & 128.44 & 107.82 & 103.47 & 129.90 & 109.23 & 114.66 & 153.66 & 134.71 \\ 
    0.99 & 0.94 & 0.99 & 0.99 & 0.99 & 0.99 & 0.99 & 0.90 & 0.99 \\
      \bottomrule
    \end{tabular}
   }
\end{table}

\begin{table}[]
    \centering
    \caption{\footnotesize Normalized rewards and prediction accuracy of SCT against unseen strategies.}
    \label{tab:mem-testing-result}
    \footnotesize
    \resizebox{\columnwidth}{!}{%
    \begin{tabular}{ccccccccccc}
    \toprule

     & ${\textsc{mem}_1}$ & ${\textsc{mem}_2}$ & ${\textsc{mem}_3} $&$ {\textsc{mem}_4}$ & ${\textsc{mem}_5} $&$ {\textsc{mem}_6}$ & ${\textsc{mem}_7}$ &$ {\textsc{mem}_8}$ & ${\textsc{mem}_9}$ & ${\textsc{mem}_{10} }$\\ 
    \midrule
    \textbf{reward} & 101.65 & 106.96 & 236.87 & 135.07 & 135.07 & 135.88 & 175.60 & 175.60 & 130.31 & 437.77 \\ 
    \textbf{accuracy} & 0.96 & 0.96 & 0.96  & 0.96  & 0.56 & 0.84 & 0.96 & 0.95 & 0.95  & 0.99 \\ \bottomrule
    \end{tabular}%
    }
\end{table}

\noindent\textbf{SCT Equilibrium Behavior.} We now discuss some interesting behavior patterns displayed by SCT when testing against three representative strategies: \textbf{all\_d} (always defect), \textbf{all\_c} (always cooperate), and the SCE strategy \textbf{pavlov}. The first observation is that when playing against \textbf{all\_d}, after the misbelief in the first step, SCT quickly realizes the opponent's \textbf{all\_d} strategy (i.e., 0.99 accuracy) and plays \textbf{all\_d} as well, reaching a NE equilibrium $(D,D)$ in IPD (which is also a SCE). When playing against \textbf{pavlov}, SCT first chooses to cooperate for a few rounds. Since \textbf{pavlov} also opts for cooperation in these rounds, SCT starts to play defection to exploit the opponent for one round, misbelieving the unconditional cooperation from the opponent. Yet, such exploitation is penalized by the \textbf{pavlov} opponent, who also switches to defection one round later. Learning its lesson, SCT falls back to cooperation thereafter. Finally, the players reach another NE (SCE) $(C,C)$, receiving higher rewards than in the first scenario. Another interesting observation is that when facing \textbf{all\_c} opponent, SCT starts to exploit the opponent by playing defection. Yet, it also plays cooperation occasionally, for which we speculate that SCT memorizes some periodic plays in the offline dataset. In summary, belief conditioning and self-confirming loss equip the vanilla transformer with greater adaptability to sophisticated opponents. 
\begin{table}[]
    \centering
    \caption{\footnotesize SCT v.s. Pavlov. After first a few rounds of cooperation, SCT starts to \textcolor{Dandelion}{exploit} Pavlov. Yet, Pavlov will \textcolor{RedOrange}{penalize} such exploitation, forcing SCT to cooperate and leading to NE plays. }
    \footnotesize
\begin{tabular}{ccccccccc}
\toprule
    {pavlov}& $\cdots$ & C & C & C & \cellcolor{RedOrange!60}D & C &  C  &$\cdots$ \\
     {SCT}  & $\cdots$ & C & C & \cellcolor{Dandelion!80}D & D & C&   C  &$\cdots$ \\
\bottomrule
\end{tabular}
\end{table}

\section{Conclusion}
Inspired by the self-confirming equilibrium (SCE), this work has developed a novel auto-regressive training paradigm for decision transformers in offline MARL tasks. The key operation of the proposed self-confirming transformer (SCT) is belief conditioning: the transformer first generates a fictitious token representing its inference about the opponent's action, which is then fed back to itself to generate its own action. The SCE-motivated loss consists of belief consistency loss and best response loss,  mandating that the agent behave optimally under the correct belief. Experimental results in multi-particle environments demonstrate SCT's superior performance against nonstationary opponents unseen in the training. Moreover, when deployed in the iterated prisoner's dilemma, SCT indeed displays equilibrium behaviors as instructed by the self-confirming loss. 

One of the most pressing future works is to investigate the interplay between return and belief conditioning. Our experiments employ the grid search to find the optimal return conditioning for all transformer models. Since the two conditionings are essentially the HIM technique \cite{furuta21generalDT}, it would be helpful if the transformer could also self-adapt return conditioning, together with belief.    


\bibliographystyle{ACM-Reference-Format} 
\bibliography{sample}


\end{document}